\documentclass[11pt,a4paper]{article}
\usepackage[nohyperref]{naaclhlt2019}
\usepackage{times}
\usepackage{latexsym}
\usepackage{graphicx}  
\usepackage{url}
\usepackage{subfig}
\usepackage{multirow}
\usepackage{amsmath}
\usepackage{caption}
\usepackage{mathabx}
\usepackage{enumitem}
\newlist{myitemize}{itemize}{1}
\setlist[myitemize,1]{label=\textbullet,leftmargin=10pt}

\aclfinalcopy 

\title{Train One Get One Free: Partially Supervised Neural Network for \\ Bug Report Duplicate Detection and Clustering}

\author{Lahari Poddar$^{1,2}$ \\ {\bf Luis Marujo$^2$}
 \And Leonardo Neves$^2$ \\ {\bf Sergey Tulyakov$^2$} \\\\
 $^1$School of Computing, National University of Singapore \\
  {\tt lahari@comp.nus.edu.sg} 
  \\
  $^2$Snap Research \\
  {\tt \{lneves, willb, luis.marujo, stulyakov, pradeep.karuturi\}@snap.com}
 \And   William Brendel$^2$ \\  {\bf Pradeep Karuturi$^2$}}

\begin{document}
\maketitle
\begin{abstract}
Tracking user reported bugs requires considerable engineering effort in going through many repetitive reports and assigning them to the correct teams.
This paper proposes a neural architecture that can jointly (1) detect if two bug reports are duplicates, and (2) aggregate them into latent topics. 
Leveraging the assumption that learning the topic of a bug is a sub-task for detecting duplicates, we design a loss function that can jointly perform both tasks but needs supervision for only duplicate classification, achieving topic clustering in an unsupervised fashion. 
We use a two-step attention module that uses self-attention for topic clustering and conditional attention for duplicate detection.
We study the characteristics of two types of real world datasets that have been marked for duplicate bugs by engineers and by non-technical annotators.
The results demonstrate that our model not only can outperform state-of-the-art methods for duplicate classification on both cases, but can also learn meaningful latent clusters without additional supervision.

\end{abstract}
\section{Introduction}
User feedback is a key part of the development and improvement of software products. 
Each piece of feedback needs to be manually reviewed and assigned to the correct engineer responsible for maintaining the feature mentioned in the report.
On online platforms with millions of users, different users tend to report the same issue, yielding a large number of duplicate reports. Sorting through these massive volumes of bug reports incur a significant amount of engineering time and cost. 
Additionally, these services are constantly releasing new product features. Therefore, we cannot rely on static annotated data for product feature classification because they rapidly become outdated.
This motivates us to develop a framework that can automatically identify duplicate bug reports and cluster them without requiring additional labels. 

Previous research in the software engineering domain has addressed duplicates detection and product feature identification as two separate problems framed as independent fully-supervised classification tasks \cite{nguyen2012duplicate, budhiraja2018lwe,jonsson2016automated,mani2018deeptriage}.
However, we observe that users generally report issues when accessing certain features of a product, e.g., `\textit{app crashed when opening camera}', `\textit{chat won't load}' and so on (here, \texttt{camera} and \texttt{chat} are the product features, respectively). Hence, two reports should \emph{at least} discuss the same feature to be considered as duplicates. Therefore, we hypothesize that determining the feature discussed in a report is a \emph{sub-task} of detecting whether or not a report is a duplicate to another one.

Inspired by the effectiveness of Siamese architectures for modeling pairs of texts \cite{tan2015lstm, bowman2015large}, we use a shared Recurrent Neural Network (RNN) to encode the two reports. We note that the latent vectors, learned by an RNN for a sequence of words, encode a multitude of semantic information; all of which may not be necessary to understand the topic of the report. 
We decompose the latent semantic vectors in order to distill only the topical information in a few designated dimensions.
This allows us to perform the sub-task of feature-based clustering using only a subset of dimensions, and use the complete vector for the duplicate classification task. 
We propose a partially supervised learning framework that uses the label for duplicity through a similarity loss on the designated topic dimensions of the latent representation to learn topic clusters. 

We use a two-step attention module, since the same words are often not crucial for both tasks. We first learn a self-attention for topic similarity modeling and
learn a conditional attention using a memory vector for duplicate classification. 
To summarize, we present a systematic study of a classic problem in the software industry. The work has three major contributions. 

\begin{itemize}
    \item We propose a neural model for multi-task learning that requires supervision for only one of the tasks.
    \item The model uses semantic space decomposition and a hierarchical-conditional attention module to perform the tasks of duplicate detection and topic based clustering.
    \item We present the challenges we faced during our experience obtaining labels from non-technical annotators, and conduct extensive experiments on both engineer labeled and non-technical labeled datasets.
\end{itemize} 
\section{Related Work}
The software engineering community has conducted some research work for  detecting duplicate bugs. In \newcite{minh2014approach},  a combination of n-gram features and cluster shrinkage algorithm is used for duplicate classification, whereas,  in \newcite{sun2011towards} the BM-25 based scoring has been used akin to information retrieval engines. A combination of tf-idf and topics learned by LDA have also been used \cite{nguyen2012duplicate,budhiraja2018lwe}.
Recently, word embeddings have been used to compute similarity of two reports\cite{yang2016combining, budhiraja2018dwen}. However, in these approaches, the sequence information of a natural language sentence is not captured.

Among the NLP community, the closest line of applicable work are the generic approaches of textual similarity \cite{cer2017semeval, wang2017bilateral, neculoiu2016learning, yin2016abcnn}. They predominantly employ Siamese architecture \cite{mueller2016siamese, severyn2015learning} and more recently, attention mechanism \cite{wang2017bilateral, shen2018sentiment,wang2018one,tran2018multihop}. Inspired by these approaches, we propose a solution for duplicate detection while utilizing the partial supervision to achieve a sub-task of clustering \textit{for free}.

Solving owner attribution of a report has been approached using feature-based methods \cite{jonsson2016automated, xuan2012developer} and very recently using deep learning solutions \cite{mani2018deeptriage}. However, the connection between these two problems have not been explored and they require supervision for product feature identification.
\section{Annotation Challenges}
\label{annotation}
We consider a dataset consisting of user reported bugs collected from the Snapchat app.
The bugs have been submitted by beta-testers using a bug-tracking functionality named Shake2Report(S2R) within the app. In S2R a user can submit a small textual description of the bug and attach a screenshot of the app while experiencing the issue. In this work, we only consider the textual descriptions of the reports.

Following previous work of duplicate bug tracking \cite{lazar2014generating}, we first studied $500$ pairs of reports marked as duplicates by engineers.
However, these pose several challenges. Firstly, we realized that engineers often needed additional meta-data such as stack-traces and timestamps to accurately determine whether or not a pair of reports referred to the same problem. On the other hand, such additional meta-data cannot be made available to external crowd-sourced annotators due to legal and user privacy restrictions. Finally, the total amount of reports was far beyond what an engineering team could tackle, hence the need for our classification system, which helps scaling the bug report load while maintaining the annotation quality.

Our task is to automatically classify report pairs, and to add engineers to the loop only when pairs are textually ambiguous. We asked non-expert annotators to label the pairs only according to their semantic similarity (no metadata). The annotators had a high agreement among themselves with a Krippendorff's alpha \cite{krippendorff} score of $0.78$. After adding engineers to the annotation pool, the score dropped to $0.58$. 
To understand this
disparity, consider the
pairs:
\begin{myitemize}
\itemsep -3pt
    \item \textit{``Crashed when posting story from memories''} and \textit{``Crashed while exporting from memories''}
    \item \textit{``Stuck on typing notification''}  and \textit{``Did not get notification for this group chat''}
\end{myitemize}
\vspace{-5pt}
Although not marked as duplicates by non-experts, they  referred to the same underlying problem from the engineering side.
While out-of-scope for this paper, we plan to conduct a dedicated study to explore these differences in future work.

As mentioned in the introduction, bug$\blacktriangleright\,$team assignments become rapidly outdated as the app product features (and therefore the teams working on them) are constantly evolving.
Also, non-expert annotators cannot be assumed to have any knowledge of these assignments, so we treat bug$\blacktriangleright\,$team assignments as an unsupervised task.
\section{Proposed Method}

\begin{figure*}[htbp]
    \centering
    \includegraphics[width=0.95\linewidth]{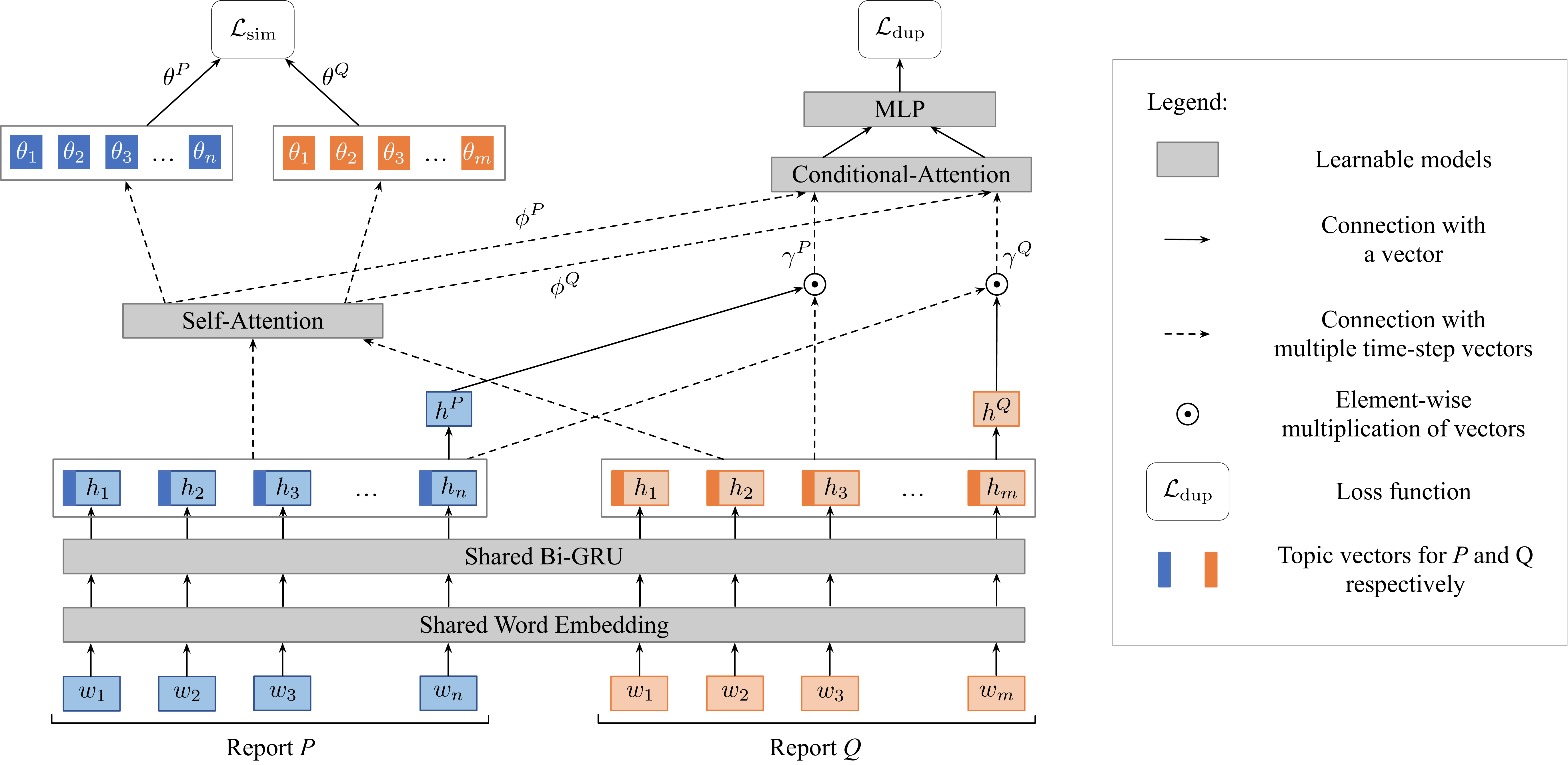}
    \caption{\textbf{The proposed partially supervised learning framework.} The framework takes two user reports $P$ and $Q$ as inputs. The left branch performs topic clustering with self-attention mechanism using topic dimensions. The right branch performs duplicates detection with conditional attention taking a pair of encoded reports as inputs.}
    \label{arch}
\end{figure*}

We now describe the proposed learning framework as shown in Figure \ref{arch}. 
Given a pair of reports $P$ and $Q$, the system uses a single binary label $r(P,Q)$ (1~if the reports are duplicate, 0~otherwise) to supervise both tasks: duplicate classification and topic similarity modeling.

\subsection{Text Encoder}
\label{method:text_encoder}
This component takes a bug report represented as a sequence of words $\{ w_1, w_2, \cdots , w_n\}$ as input and encodes it to latent vectors $\{ \mathbf{h}_1, \mathbf{h}_2, \cdots , \mathbf{h}_n\}$.

We first use word embeddings to transform all words in a text into finite $d$-dimensional vectors $\{\mathbf{v}_1, \mathbf{v}_2, \cdots ,\mathbf{v}_n\}$.
The vectors are then fed to a Gated Recurrent Unit (GRU) layer \cite{chung2014empirical}.
We use bi-directional GRU units to encode the context information around a word. For a word $w_i$, outputs from the forward and backward GRUs are latent vectors of dimension $g$ and denoted as $\mathbf{h}^f_i$ and $\mathbf{h}^b_i$, respectively.

We concatenate these two vectors to form a single latent representation of the word as, $\mathbf{h}_i = \mathbf{h}^f_i \oplus \mathbf{h}^b_i$ .

\subsection{Semantic Space Decomposition}
\label{method:decompose}
A word's encoded representation $\mathbf{h}_i$ has various information intertwined in an abstract way within the vector dimensions. It is difficult to interpret individual dimensions and meaningfully use a subset of dimensions for a different sub-task. 

We aim to distill the coarser topical information into designated dimensions within the vector, storing other finer pieces of information in remaining dimensions. Such disentanglement allows us to do the topic similarity modeling using only the designated dimensions and ignore the rest.

To this end, we force the network to learn topical information about each word in the first $k$ dimensions of its latent vector space. 
For a bi-directional GRU, we have two encoded representations for the $i^{th}$ word  from the forward and backward passes i.e., $\mathbf{h}^f_i$, and $\mathbf{h}^b_i$. 
We define the topic vector of a word $i$ as,
\begin{align}
    \boldsymbol{\theta}_i = \mathbf{h}^f_i[1:k] \oplus \mathbf{h}^b_i[1:k]
\end{align}
\noindent where $\mathbf{h}^f_i[1:k]$ and $\mathbf{h}^b_i[1:k]$ denote the first $k$ dimensions in the GRU encoded latent vectors from the forward and backward passes respectively.

\subsection{Topic Similarity Modeling}
\label{method:topic}
We aggregate the topic vectors of the constituent words to represent the topic of a report, and use a self-attention layer to learn the weights of the words important in determining the topic:
\begin{align}
    &\mathbf{z}_i = \mathrm{tanh}(\mathbf{W}\cdot\mathbf{h}_i + \mathbf{b}_i)\\
    &\boldsymbol{\alpha}_i = \frac{\mathrm{exp}(\mathbf{z}_i)}{\sum_i \mathrm{exp}(\mathbf{z}_i)},\\
    &\boldsymbol{\theta} = \sum_{i=1}^n \boldsymbol{\alpha}_i \cdot \boldsymbol{\theta}_i.
\end{align}
\noindent where $\boldsymbol{\alpha}_i$ is the weight for word $w_i$ and $\boldsymbol{\theta}$ is the topic vector of the report.
Note that only the designated topical dimensions of the words contribute to representing the topic vector of the report. 

In order to learn the semantic space of topic vectors, we need to constrain them in such a way that if two reports ($P$ and $Q$) have similar topics (e.g., both are talking about \textit{camera}), their $\boldsymbol{\theta}^P$ and $\boldsymbol{\theta}^Q$ values should be closer and vice-versa.
Since only the ground-truth label for duplicity is available, we consider this signal as \textit{partially} observed for the topic modeling task. Following our hypothesis that determining the topic of a report is a sub-task for duplicate detection, there can be three possible cases for a pair of reports:

\smallskip
\noindent \textbf{Case 1:} Duplicates from the same topic;

\noindent \textbf{Case 2:} Non-duplicates from different topics;

\noindent \textbf{Case 3:} Non-duplicates from the same topic.
\smallskip

Since for duplicate tickets ($P, Q$) their topics are bound to be same, we reduce the distance between $\boldsymbol{\theta}^P$, and $\boldsymbol{\theta}^{Q}$ in our loss. 
However, non-duplicate pairs may or may not be from the same topic (Case 2, and 3).

We do not include Case 3 in the loss, as inferring it is difficult without explicit labels. We assume that if a pair of non-duplicate reports have no word overlap (apart from stopwords), then only they belong to different topics i.e., Case 2. For such pairs we wish to increase the distances between their topic vectors. We imbue this principle in the network through the following loss function:
\begin{align}
    \mathcal{L}_\mathrm{sim} & = r(P, Q) \cdot S_\mathrm{C}(\boldsymbol{\theta}^P, \boldsymbol{\theta}^Q)\\\nonumber 
    & - (1-r(P,Q)) \cdot S_\mathrm{C} (\boldsymbol{\theta}^P, \boldsymbol{\theta}^Q) 
\end{align}
 where $r(P, Q)$ is the ground-truth duplicity label ($0$ or $1$), $S_\mathrm{C}$ is cosine similarity. A careful reader might observe that the function minimizes distance of duplicates pairs ($r(P, Q) = 1$), while exhibiting the exact opposite behaviour for non-duplicates ($r(P, Q) = 0$). To account for the skewed distribution of duplicate vs. non-duplicates, we normalize the loss using proportional class weights. 

\subsection{Duplicate Classification}
\label{method:dedup}
The final component of the network considers the complete latent vectors of the words for the task of duplicate classification. We use another attention layer for this component to allow the network to focus on important words. However, it is not necessary that the same words will be important for both the tasks. For example, consider the following pair of tickets:
\textit{``Can't import these pics from camera roll to memories''} and \textit{``No pics in memory''}.

In order to determine the product feature, the self attention module needs to focus on the words \textit{memories} and \textit{memory} in the reports, respectively. However, to decide whether they are duplicates or not, the second attention module needs to focus on the specific error, i.e. \textit{Can't import pics} and \textit{No pics} in the two reports respectively.
For report $P$, we create a memory vector $\boldsymbol{\phi}^P$ to guide this attention layer using the hidden representations of the words as well the previously learned self-attention weights for topic modeling.
\begin{align}
    \boldsymbol{\phi}^P = \sum_i \boldsymbol{\alpha}_i^P \cdot \mathbf{h}_i^P + \sum_i \mathbf{h}_i^P
\end{align}
We also note that unlike the self-attention used for topic similarity modeling, for duplicate classification we need to use a form of conditional attention. For determining whether a report ($P$) is duplicate to another one ($Q$), the words that are important in $P$ are dependent on the words in $Q$. Therefore, for each word position $i$ in $P$ we compute its relevance to the report $Q$ as,
\begin{align}
    \boldsymbol{\gamma}^{P}_i = \mathbf{h}_i^P \odot \mathbf{h}^Q
\end{align}

where $\mathbf{h}_i^P$ is the hidden state of the $i^{th}$ word in report $P$, $\odot$ denotes element-wise multiplication, and $\mathbf{h}^Q$ is the representation of report $Q$ obtained using averaging the hidden states of all words in $Q$ i.e. $\mathbf{h}^Q = \mathrm{Avg}(\mathbf{h}_1, \mathbf{h}_2, \cdots, \mathbf{h}_m)$, $m$ is the number of words in $Q$. 

We now use this conditional representation $\boldsymbol{\gamma}^{P}_i$ and the memory vector $\boldsymbol{\phi}^P$ to learn the conditional attention weights ($\boldsymbol{\beta}$) and compute the weighted representation $\mathbf{c}^P$ as,
\begin{align}
    \mathbf{c}_i^P & = \mathrm{tanh}(\mathbf{W}_1 \cdot \boldsymbol{\gamma}^{P}_i) \cdot \mathrm{tanh}(\mathbf{W}_2\cdot\boldsymbol{\phi}^P)\\
    \boldsymbol{\beta}^{P}_i & = \frac{\mathrm{exp}(\mathbf{c}_i^P)}{\sum_i \mathrm{exp}(\mathbf{c}_i^P)}\\
    \mathbf{c}^P & = \sum_{i=1}^n \boldsymbol{\beta}^{P}_i \cdot \mathbf{c}_i^P
\end{align}
We concatenate the weighted representations of $P$ and $Q$ as $\mathbf{c}^{PQ} = \mathbf{c}^P \oplus \mathbf{c}^Q$. We pass this concatenated vector through a Multi-Layer Perceptron, the final layer of which outputs a prediction $\hat{r}(P,Q)$ whether they are duplicates or not. We use binary cross-entropy loss to train this objective.
\begin{align}
    \mathcal{L}_\mathrm{dup} = H(r(PQ),\hat{r}(PQ)),
\end{align}
where $H$ is binary cross-entropy. The overall loss for the whole network is a weighted combination of the two losses.
\begin{align}
    \mathcal{L} = \lambda\mathcal{L}_\mathrm{sim} + (1-\lambda) \mathcal{L}_\mathrm{dup}
\end{align}

In our experiments, we use $\lambda = 0.5$ but it can be varied depending on the importance of the tasks.
We optimize the network using Adam optimizer and train in an end-to-end fashion through backpropagation.
\section{Evaluation}

\subsection{Dataset}
We experiment on four real-world datasets.

 \textbf{(1) Snap S2R }: Composed of bugs reported through an in-app tool. 
 The data was labeled as duplicate or not by non-technical annotators (described in section \ref{annotation}). 

 \textbf{(2-4). Open Source Projects} \footnote{\url{https://github.com/logpai/bugrepo}} : Repository of bugs submitted for the open-source projects of \textbf{Eclipse Platform}, \textbf{Mozilla Firefox} and \textbf{Eclipse JDT}. Each report consists of a short title and a longer description which often describes the steps to reproduce the issue. We only consider the title of the report in our experiments. 
 The reported bugs have been marked for duplicates by engineers while resolving them. 


We augment the S2R data with randomly sampled negative pairs to resemble a positive to negative ratio that is estimated in production.
For the Eclipse and Firefox datasets, we randomly sample negative pairs, keeping the positive to negative ratio close to what was previously mentioned \cite{lazar2014generating}.  Table \ref{dat_stats} shows the statistics of the four datasets. 
These datasets were selected to capture different training sizes and variations  of vocabulary used. 

\begin{table}[htbp]
\centering
\resizebox{\linewidth}{!}{%
\begin{tabular}{l|lllll}
\hline
Dataset & \#pairs  & \#vocab & \%dups & \#reports & \multicolumn{1}{l}{\begin{tabular}[l]{@{}l@{}}avg \#words\\/report\end{tabular}} \\ \hline
Snap S2R & 66,945 &  7763  & 9\% & 17,255 & 14.6\\ 
Eclipse Platform & 170,312  & 29,702 & 12\% & 83,608 & 7.9\\
Eclipse JDT & 90,592  & 17,228 & 14\% & 44,670 & 8.1\\ 
Firefox & 231,628 & 34,590 &  26\% & 113,262 & 10.0\\ \hline
\end{tabular}
}
\caption{Statistics of the datasets used}
\label{dat_stats}
\end{table}

\subsection{Experimental Settings}
 We initialize the words using  pre-trained Glove embeddings \cite{pennington2014glove} of 300 dimensions but tune it during training to capture the intrinsic features of the specific task and dataset at hand. For the GRU layers we use 150 dimensions and the first 20 dimensions of them are used for topic representation. For the MLP in duplicate classification we use $2$ layers of fully connected layers of $100$ hidden neurons each, with $relu$ activation and $20\%$ dropout rate. For fair comparison we use the same number of parameters in all baselines. The learning rate is set to $0.003$ and a batch size of $128$ samples is used while training all the models.

\subsection{Evaluating Duplicate Classification}

We first start evaluating the proposed approach for detecting duplicate bug reports. We compare with the following six baselines:
\begin{itemize}
    \item \textbf{Logistic Regression} uses a bag of $n$-grams representation with $n$ ranging from one to three words. Tf-idf scores are used for feature weighting. 
\item \textbf{Siamese CNN} \cite{severyn2015learning} encodes the texts with a shared convolution network followed by an MLP for classification.
\item \textbf{Siamese Bi-GRU} uses a shared bidirectional GRU for encoding text. 
\item \textbf{Siamese Bi-GRU with Attention} uses an attention layer on top of the GRU outputs to encode the texts
\item \textbf{DWEN} \cite{budhiraja2018dwen} is the state-of-the-art deep learning approach for duplicate bug detection using word embeddings.
\item \textbf{BiMPM} \cite{wang2017bilateral} is the state-of-the-art sentence similarity modeling that uses multi-perspective symmetric matching for a pair of texts.
\end{itemize}

\begin{table*}[ht]
\centering
\resizebox{\linewidth}{!}{%
\begin{tabular}{|l|lll|lll|lll|lll|}
\hline
\multirow{2}{*}{Method} & \multicolumn{3}{c|}{Snap S2R} & \multicolumn{3}{c|}{Eclipse Platform} & \multicolumn{3}{c|}{FireFox}&\multicolumn{3}{c|}{Eclipse JDT} \\ \cline{2-13} 
 & P & R & F1 & P & R & F1 & P & R & F1 & P & R & F1 \\ \hline
Logistic Regression & 0.67 & 0.63 & 0.65 &  0.71 & 0.88 & 0.78 & 0.92 & 0.95 & 0.94 & 0.76 & 0.88  &  0.81  \\ \hline
Siamese CNN \shortcite{severyn2015learning} & 0.67 & 0.63 & 0.65 & 0.81 & 0.81 & 0.81 & 0.93  & 0.94 & 0.93 & 0.79 & 0.79 & 0.79 \\ \hline
Siamese Bi-GRU & \bf{0.76} & 0.61 & 0.68 & \bf{0.86} & 0.84 & 0.85 & \bf{0.95} & 0.93 & 0.94 & 0.85 & 0.84 & 0.84 \\ \hline
Siamese Bi-GRU w Att & \bf{0.76} & 0.62 & 0.68 & \bf{0.86} &  0.86 &  0.86 & \bf{0.95} & 0.94 & 0.94 & 0.84 & 0.84 & 0.84 \\ \hline
DWEN \shortcite{budhiraja2018dwen} & 0.69 & 0.55 & 0.62 & 0.83 & 0.74 & 0.78 & 0.93 & 0.92 &0.93 & 0.81 & 0.71 & 0.76 \\ \hline
BiMPM \shortcite{wang2017bilateral} & 0.73 & 0.65 & 0.69 & \bf{0.86} & 0.82  & 0.84 & 0.94  & 0.92 & 0.93 & 0.85 & 0.79 & 0.82 \\ \hline
Our Method & 0.73 & \bf{0.67*} & \bf{0.70} & 0.84 & \bf{0.91*} & \bf{0.87*} & 0.94 & \bf{0.96*} & \bf{0.95*} & \bf{0.86} & \bf{0.90*} & \bf{0.88*} \\ \hline
\end{tabular}
}
\caption{Comparison of different methods for duplicate classification task on multiple datasets. * denotes statistical significance with the runner up for $p$\textit{-}$value <0.01$}
\label{dupAcc}
\end{table*}

We report results after averaging five independent trials using $80\%$ data for training, $10\%$ for validation, and $10\%$ for testing.
Due to the imbalanced  class distribution, we use precision, recall and F1-score of the positive class as evaluation metrics.

From the results in Table~\ref{dupAcc} we see that our proposed method largely outperforms all other models in terms of recall and in most cases on F1 score as well. Using a sequence encoder like GRU boosts the performance significantly compared to using n-grams as in Logistic Regression or CNN. 
We note that the overall scores on Eclipse and Firefox datasets are much higher than the scores on Snap S2R data. One of the reasons could be the difference in data size where the Snap S2R dataset is much smaller in size compared to others. Noisy labels from non-technical annotators could also be a potential reason. Manually examining the posts we also note that the bug reports submitted for the open projects are often by engineers, who use concrete technical terms to describe the problem. In contrast, for the Snap S2R dataset, the reports are submitted by end-users using free text with non-technical terms that make it harder for a machine learning model to disambiguate. 

\subsection{Evaluating Clusters}

A major advantage of our model is its ability to perform clustering of the bug reports to aid in ownership attribution.
In this set of experiments, we show that our model is able to learn a semantic space so that nearby reports in the space indeed belong to the same product feature.

We use the topic vector ($ \theta $) of a report learned by our model and run an off-the-shelf clustering algorithm K-means\footnote{\url{https://scikit-learn.org/stable/modules/generated/sklearn.cluster.KMeans.html}}. Ideally, there should be a one-to-one mapping between the obtained clusters and the actual project features. 
For comparison, we use one of the most popular unsupervised topic model, LDA~\cite{blei2003latent} to learn latent topics from the bug reports. 
For implementation of LDA we use the Gensim library\footnote{\url{https://radimrehurek.com/gensim/models/ldamodel.html}}. We perform standard pre-processing steps, and remove stop-words before training the LDA model.
We evaluate the learned latent clusters by comparing them to the ground truth project features,
as have been marked by users while submitting reports for S2R dataset.
For fair comparison, we use the same ($20$) number of clusters for both methods, which is close to the actual number of product features.

\begin{figure}[ht]
    \centering
    \subfloat[Our Method]{\includegraphics[scale=0.1,width=0.7\linewidth]{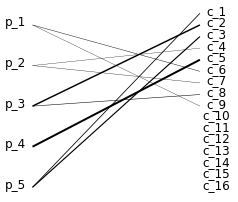}} 
    \\ 
    \subfloat[LDA]{\includegraphics[scale=0.1,width=0.7\linewidth]{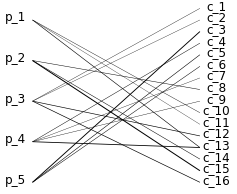}}
    \caption{Mapping between the learned latent clusters and top-$5$ product features for Snap S2R dataset.}
    \label{clustermaps}
\end{figure}

Figure \ref{clustermaps} shows the mapping between learned latent clusters and top-$5$ ground truth product features for the proposed model and LDA. 
To reduce the effect of noisy labels, we consider a cluster-product feature mapping only if more than $10$ reports about a product feature are assigned to a cluster. 
The width of the connecting edge is proportional to the number of times a cluster-product feature association is observed.

Firstly, we observe that LDA requires more clusters to represent the same number of features. This demonstrates that LDA over-clusters, failing to identify reports that talk about the same product feature and assigns them to different clusters. Secondly, some clusters in LDA (e.g., $c\_{13}$) are connected to multiple product features denoting impurity of the cluster. Finally, for most features, their mappings to a cluster are much stronger in our model as demonstrated by the width of the edges in the graph. 
This shows that with partial supervision from duplicity labels, we are able to learn clusters that better correspond to ground truth product features.

\begin{table}[htbp]
\centering
\resizebox{\linewidth}{!}{%
\begin{tabular}{|l|l|l|l|l|}
\hline
Cluster 1 & Cluster 2 & Cluster 3 & Cluster 4 & Cluster 5 \\ \hline
video   & load & lens& app & specs\\
audio   & long & lenses& map & spec \\
videos  & story & carousel& viewing & importing\\
playback & time & preview & crashed & v2\\
sync  & taking & applied & navigating & unpaired\\ \hline
\end{tabular}
}   
\caption{Top Words according to their tf-idf scores from few clusters learned by our model.}
\label{topWords}
\end{table}

Table \ref{topWords} shows top words according to their tf-idf scores appearing in bug reports assigned to a cluster. The learned clusters are coherent and represent specific product features. For example, \textit{Cluster 1} talks about issues regarding \texttt{video} or \texttt{audio}, while \textit{Cluster 2} is mostly about issues related to loading of a \texttt{story} (a feature in the app). In \textit{Cluster 3} reports related to various \texttt{lenses} and \texttt{filters} are placed, while \textit{Cluster 4} seems to be about \texttt{maps} and \texttt{navigation} features. It is also interesting to note that generic issues like \textit{`crashed'} appear in some of the clusters. Although one could identify most of the bugs as crashes, this indicates that users have a specific vocabulary when referring to a particular product feature. 

\subsection{Analyzing Attention Weights}
Finally, we present case studies of the attention weights learned by our proposed model. As we have two-steps of attention layers optimized by the two objectives, they can learn to focus on words that are important for the specific tasks. 

\begin{figure}[ht]
    \subfloat[bug report: \textit{totally black pic for \textbf{tile} story} \label{att_sample1}]{\includegraphics[trim=2pt 2pt 2pt 20pt, clip,width=\linewidth]{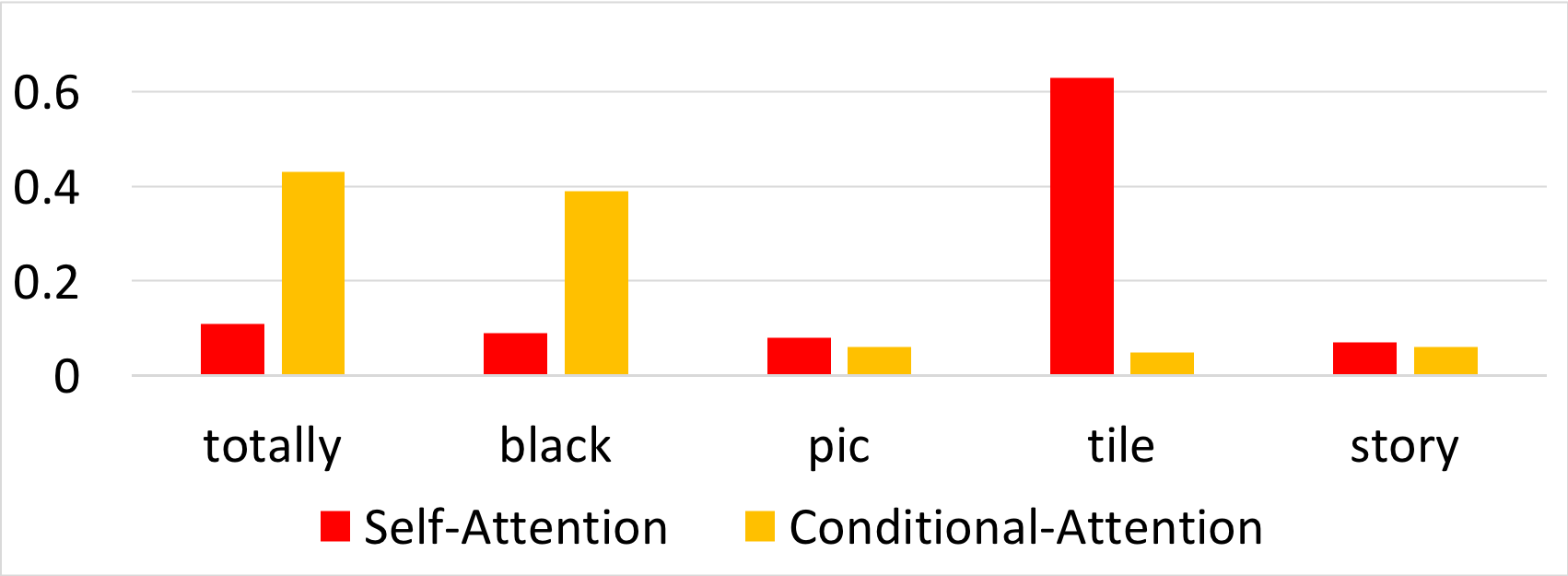}
    } \\ 
    \subfloat[bug report: \textit{pic importing despite \textbf{specs} disconnected} \label{att_sample2}]{\includegraphics[trim=2pt 2pt 2pt 10pt, clip,width=\linewidth]{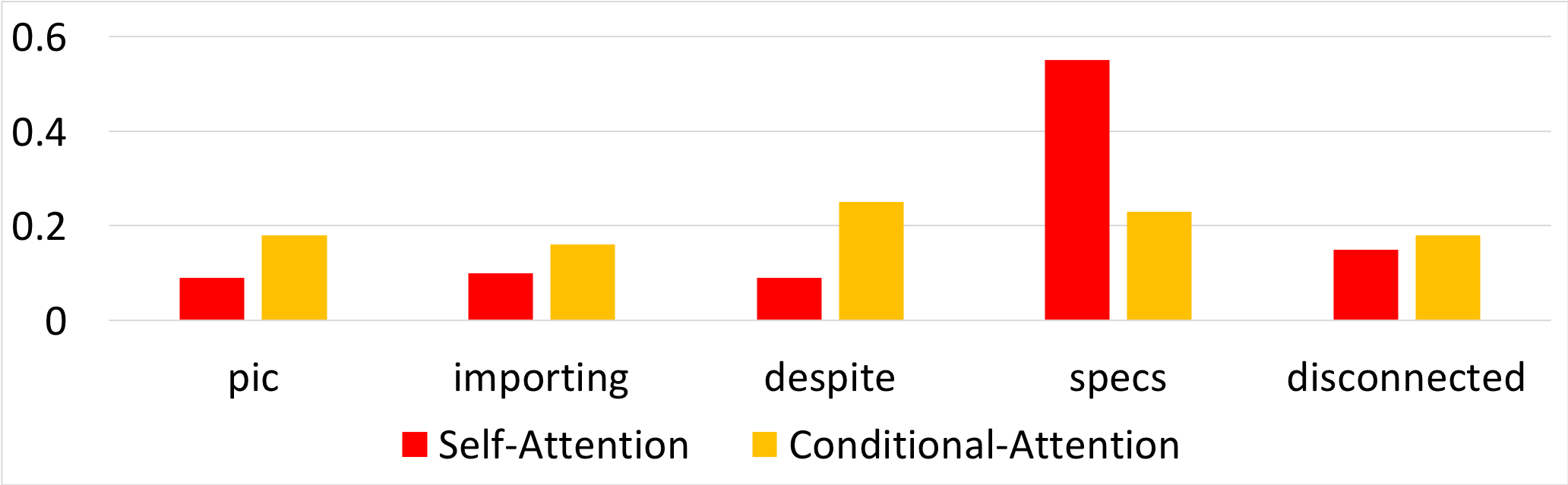}} 
    
    \caption{Visualization of the attention weights learned by the two attention modules on few sample bug reports}
    \label{attWeights}
\end{figure}

Figure \ref{attWeights} shows the attention weights for two sample bug reports from Snap S2R dataset. We observe that the two attention modules tend to focus on different parts of the text. The self-attention gives more weight to the words that determine the overall topic of the report (\textit{tile} in Figure \ref{att_sample1}, \textit{specs} in Figure \ref{att_sample2}). In contrast, the conditional-attention focuses on the set of words describing the specific issue to aid in detecting duplicates.

\section{Conclusion}
In this paper we have studied bug-tracking, which is a widespread problem in the software industry. We develop a neural architecture that can learn to classify duplicates and cluster them into meaningful latent topics without additional supervision. The architecture decomposes the latent semantic space of a word to only distill the topical information into a few designated dimensions, and uses a two-step attention module to focus on different textual parts for the two tasks. We share the challenges of annotating a user reported bug dataset with non-technical annotators, as opposed to using annotations from engineers. This is a direction we plan to further explore in future work. Experimental results on different types of datasets indicate that the proposed approach is promising compared to existing techniques for both tasks. 
Most importantly, our model's construction is generic and presents new possibilities in various domains for modeling sub-tasks for free, with partial supervision from another task. 

\newpage
\bibliography{References}
\bibliographystyle{acl_natbib}

\end{document}